# Cross-attention-based saliency inference for predicting cancer metastasis on whole slide images

Ziyu Su, Mostafa Rezapour, Usama Sajjad, Shuo Niu, Metin Nafi Gurcan, Muhammad Khalid Khan Niazi

*Abstract*— Although multiple instance learning (MIL) methods are widely used for automatic tumor detection on whole slide images (WSI), they suffer from the extreme class imbalance within the small tumor WSIs. This occurs when the tumor comprises only a few isolated cells. For early detection, it is of utmost importance that MIL algorithms can identify small tumors, even when they are less than 1% of the size of the WSI. Existing studies have attempted to address this issue using attention-based architectures and instance selection-based methodologies, but have not yielded significant improvements. This paper proposes cross-attention-based salient instance inference MIL (CASiiMIL), which involves a novel saliency-informed attention mechanism, to identify breast cancer lymph node micro-metastasis on WSIs without the need for any annotations. Apart from this new attention mechanism, we introduce a negative representation learning algorithm to facilitate the learning of saliency-informed attention weights for improved sensitivity on tumor WSIs. The proposed model outperforms the state-of-the-art MIL methods on two popular tumor metastasis detection datasets, and demonstrates great cross-center generalizability. In addition, it exhibits excellent accuracy in classifying WSIs with small tumor lesions. Moreover, we show that the proposed model has excellent interpretability attributed to the saliency-informed attention weights. We strongly believe that the proposed method will pave the way for training algorithms for early tumor detection on large datasets where acquiring fine-grained annotations is practically impossible.

## I. INTRODUCTION

DIGITAL pathology is playing an increasingly important role in cancer diagnosis and transforming how pathologists provide diagnostic information to patients and clinicians [1]. By digitizing cancer specimens as whole slide images (WSIs) with high resolution, pathologists can now view, share, and analyze them more easily and efficiently. One of the benefits of digital pathology is that it enables pathologists to seek a second opinion from other experts more quickly by sharing images. Moreover, it provides opportunities for the development of machine learning-based computer-aided diagnosis technologies [1, 2].

In recent years, deep learning has become the preferred machine learning method for analyzing WSIs due to its remarkable learning capabilities [3]. However, there are two major challenges when using deep learning models to analyze WSIs. Firstly, WSIs are often extremely large (giga-pixel size) and stored in a multi-resolution format to imitate the light microscope, making them even larger. Secondly, accurate ground truth labels describing and annotating regions of interest (e.g., lesions) are often scarce. Pathologists usually prefer to provide overall diagnostic labels (e.g., cancer or normal) rather than to take the time and effort to annotate lesions on WSIs or to draw their boundaries.

Unfortunately, MIL methods employed in detecting breast cancer micro-metastasis to the lymph nodes have not yielded satisfactory results. Identifying breast cancer metastasis to lymph nodes holds significant importance as it assists oncologists in determining the stage of breast cancer and devising treatment plans [4]. The presence of tumor cells in the lymph nodes indicates the spread of cancer beyond the breast tissue and can imply a higher stage of the disease. The size of the tumor involvement in a lymph node and the number of lymph nodes that are involved by tumor are both essential in cancer staging [5], which directly affects the treatment plan and the disease prognosis. Moreover, *an early diagnosis* of lymph node metastasis in breast cancer is vital for improving treatment outcomes and overall prognosis [6, 7]. Hence, it is imperative to develop methods that help identify micro-metastasis of breast cancer to lymph nodes at an early stage.

When dealing with WSIs that have small tumor lesions, such as micro-metastasis to the lymph nodes, the size of the tumor may comprise only a few isolated cells. For this reason, an MIL bag may contain a large number of normal instances and only a few tumor instances. This leads to a severe class imbalance during model development [8]. Some previous studies have tackled this issue by performing instance-level classification

The work was partially supported through a National Institutes of Health Trailblazer award R21EB029493 (PIs: Niazi, Segal). The research was partially funded by a National Institutes of Health grant R21 CA273665 from the NCI, with PI Gurcan leading the project. The authors take full responsibility for the content of this work, and any opinions expressed do not necessarily reflect the official views of the National Institutes of Health. (Corresponding author: Ziyu Su)

Ziyu Su, Mostafa Rezapour, Usama Sajjad, Metin Nafi Gurcan, and Muhammad Khalid Khan Niazi are with the Center for Biomedical Informatics, Wake Forest University School of Medicine, Winston-Salem, NC 27101, USA. (e-mail: zsu@wakehealth.edu, mrezapou@wakehealth.edu, usajjad@wakehealth.edu, mgurcan@wakehealth.edu, mniazi@wakehealth.edu).

Shuo Niu is with the Department of Pathology, Wake Forest University School of Medicine, Winston-Salem, NC 27101, USA (e-mail: sniu@wakehealth.edu).

42and predicting WSIs with a few high-confidence instances in the bag [2, 9]. Others have used attention-based MIL models to enable the model to focus on potential tumor instances [8, 10-13]. However, in the case of early diagnosis, the tumor lesions typically occupy a very small portion (typically less than 1%) of the WSIs, resulting in only a few tumor instances in the corresponding tumor bags. Meanwhile, the current attention mechanism tries to learn proper attention weights solely based on each single instance, without guidance from tumor-related information. Therefore, the aforementioned MIL algorithms fail to pay attention to or classify the positive instances, leading to unsatisfactory sensitivity in predicting WSIs with small tumor lesions [14]. To address this problem, several MIL studies have explored the selection of highly salient instances within MIL bags. For instance, in our previous work [14], we achieved high accuracy by pre-training an instance-level tumor detection model and using the salient instances (i.e., possible tumor instances with high-confidence) in the bags for MIL prediction. However, this approach relies on the assumption that there are some large-tumor WSIs in the dataset. Other studies combined their tumor instance detection model with gradient flow to feed salient instances to the MIL model in a learnable manner [2, 9], but detecting tumor instances solely based on slide-level label is a radical approach. As a result, these methods either achieved moderate sensitivity or required large-scale training sets.

To address the above issues, we propose a novel MIL methodology, named Cross-Attention-based Salient instance inference-MIL (CASiiMIL), that can infer possible tumor instances and mitigate the class imbalance between normal and tumor instances in an end-to-end neural network. Inspired by Transformer [15] and open-set learning methods [16-18], we propose CASii network that can automatically correlate the input instances with the representative normal instances in a more discriminative feature space (for tumor identification) and infer the salient instances dynamically by learning saliency-informed attention weights to highlight them. The contributions of this work are as follows:

- To mitigate the class-imbalance issue of small tumor WSIs encountered by existing MIL models, we propose a novel attention mechanism, named cross-attention-based salient instance inference (CASii), to learn saliency-informed attention weights for improved tumor WSI identification performance.
- We present a negative representation learning method that can learn representative normal instances to support salient instance inference.
- We introduce two instance-level loss functions to further improve the sparsity and saliency of the learned attention weights for our MIL model.

## II. RELATED WORKS

### A. MIL for WSI analysis

MIL is one of the most extensively used deep learning method for WSI analysis given its weakly supervised property. Recent MIL models can be typically divided into two categories, which are top-K instance based models [2, 9, 14, 19] and attention-based models [8, 11, 12, 20]. The top-K instance based models usually require training instance-level classifier based on the corresponding slide-level labels. However, the training of this classifier needs either large-scale WSI dataset or WSIs with macro-tumor lesions, which are not readily available. The attention-based models employ attention or self-attention [15] module to learn appropriate attention weights and aggregate the instances within the WSIs. Nevertheless, the learning of attention weight is based on slide-level labels only, making it difficult to identify tumor instances for the micro-tumor lesion cases.

### B. Open-set learning

In open-set learning, the task is to classify the categories that has been seen during the training, and identify the data from unseen categories in the meantime [17]. Typically, this is accomplished by comparing input data to example seen data [17, 18, 21]. Specifically, a previous work proposes an attention-based architecture to correlate the local regions of an input image with the local regions from a support image set [16]. Hence, their model can highlight the local image regions from the unseen categories. Inspired by open-set learning methods, the proposed method first learns a set of representative normal instances, and then utilizes a novel cross-attention mechanism to infer possible tumor instances.

## III. METHOD

### A. Datasets

Our study is based on two publicly available WSI datasets named Camelyon16 and Camelyon17 [22, 23]. They are well-known deep-learning benchmarks for the automated detection of BCLNM in hematoxylin and eosin (H&E) stained WSIs of lymph node biopsies.

Camelyon16 contains a training set with 270 WSIs and a hold-out testing set with 129 WSIs that are divided into two categories: normal and tumor. Tumor WSIs consist of both macro- and micro-metastasis WSIs, with the latter containing tumor lesions that are no larger than 0.2 mm and are more challenging to detect. In our MIL study, we perform binary classification to identify normal and tumor WSIs using only slide-level labels.

Camelyon17 dataset contains a training set with 500 WSIs and a testing set with 500 WSIs that are divided into normal and tumor categories. It is a more challenging dataset as it includes WSIs collected from five different hospitals. We utilize this dataset to assess our method's generalizability to WSIs from unseen hospitals during training. To this end, we conduct a cross-center cross-validation study based on the 500 WSIs from the Camelyon17 training set, where the source hospitals are labeled.

### B. Revisiting MIL for WSI classification

In the MIL paradigm, a WSI is first cropped into small image patches, and then via the feature extraction module, all patches are transformed into feature embeddings. Throughout this paper, we refer to a WSI and a patch as a bag and an instance, respectively. For the sake of representation, we also assumed that $X_i = \{x_{i1}, x_{i2}, ..., x_{in_i}\}$ denotes the set of all instances within the $i^{th}$ bag, where $x_{ij}$ is the $j^{th}$ instance of the $i^{th}$ bag. If an instance comes from a tumor region, then a positive label (encoded by 1) will be assigned to the instance, and otherwise, it will be labeled as negative (encoded by 0). Moreover, a bag is called positive if and only if there exists at least one positive instance in the bag. If all instances within a bag are negative, then a negative label will be assigned to the bag. Therefore, the true bag-level label, $Y$, can be defined by

$$Y = \begin{cases} 0, iff \sum_j y_j = 0, \\ 1, otherwise. \end{cases} \quad (1)$$

where $y_j \in \{0, 1\}$, for $j = 1, ..., n_i$, denotes the $j^{th}$ instance of the bag [24]. To make a bag-level decision (bag-level label prediction) for the $i^{th}$ bag, $X_i$, all instances within a bag, $\{x_{i1}, x_{i2}, ..., x_{in_i}\}$, are first transformed into instance-level embeddings. Then all instance-level embeddings are aggregated into a bag-level embedding. Finally, a classifier is employed that takes the bag-level embedding as an input and produces a bag-level label prediction as an output. The process can be formulated as:

$$\tilde{Y} = g\left(\sigma\left(f(x_{i1}), ..., f(x_{in_i})\right)\right) \quad (2)$$

where $\tilde{Y}$ is a predicted bag-level label, $f(\cdot)$ is an instance-level embedding transformation function, $\sigma(\cdot)$ is an aggregation function, and $g(\cdot)$ is a bag-level prediction classifier.

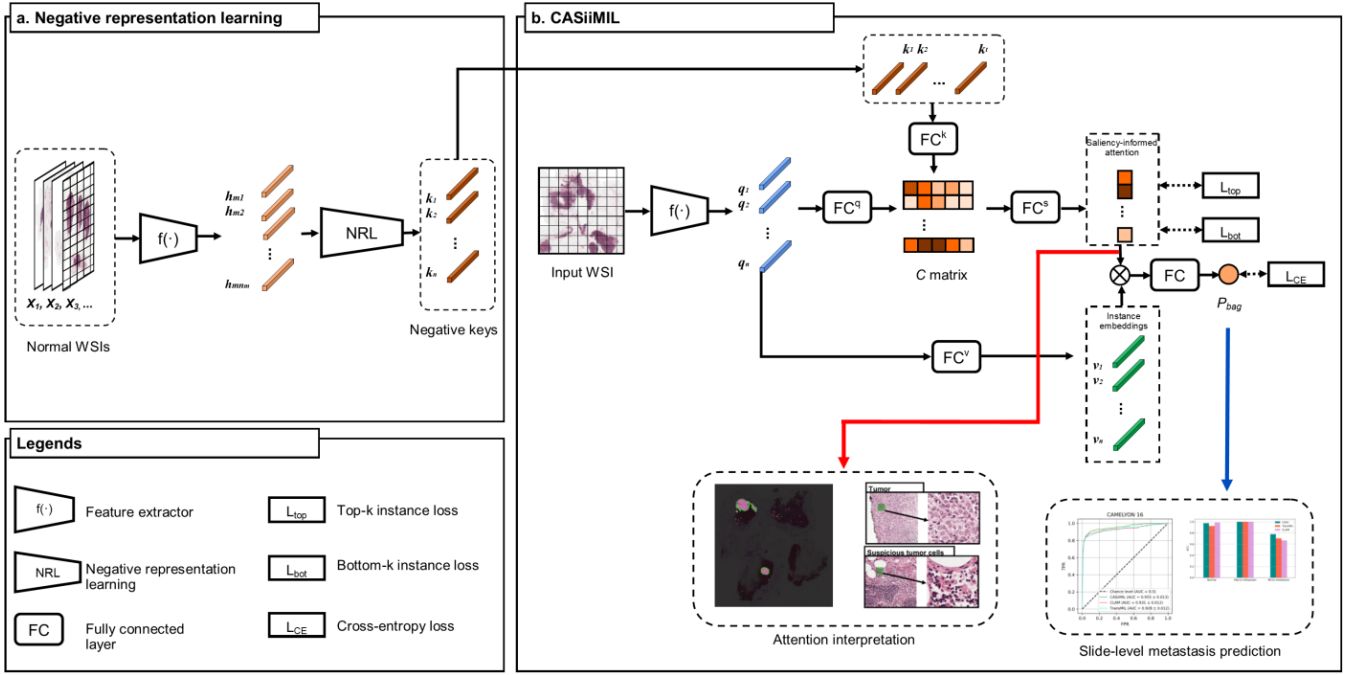

Fig 1. Overview of CASiiMIL. (a) Negative representation learning (NRL) to extract negative keys from training normal WSIs. (b) CASiiMIL for bag-level classification. Our cross-attention-based model automatically correlates the input instances and negative keys in the C matrix and then learns saliency-informed attention scores for MIL aggregation.

## C. The proposed CASiiMIL

The proposed method is composed of three main steps: (i) forming a bag by cropping an input WSI into patches and embedding the patches into feature embeddings using pretrained CNN model (see Section 3.5 and 3.6 for details); (ii) learning negative keys (i.e., representative negative instances) from normal WSIs in the training set using the proposed negative representation learning (NRL) module; (iii) bag-level classification using CASiiMIL based on an input bag and negative keys. The overview of our method is shown in Fig 1.

### 1) Negative representation learning

Given a training set that contains $P$ negative bags (normal WSIs) denoted by $X = \{X_1, X_2, ..., X_P\}$, where $X_m$ is the $m^{th}$ negative bag that contains $n_m$ negative instances (normal patches), $X_m = \{x_{m1}, x_{m2}, ..., x_{mn_m}\}$, we first apply a feature extraction neural network $f(\cdot)$ on $x_{mq} \in X_m$, for $q = 1,2, ..., n_m$ and $m = 1, 2, ..., P$, to construct feature embeddings $h_{mq} \in \mathbb{R}^D$. For $m = 1, 2, ..., P$, we then construct $A_m = [h_{m1}, h_{m2}, ..., h_{mn_m}] \in \mathbb{R}^{D \times n_m}$, apply CUR decomposition [25] on $A_m$, and export a submatrix $\tilde{A}_{b_m} \in \mathbb{R}^{D \times t_m}$. The columns of $\tilde{A}_{b_m}$, which are a subset of columns of $A_m$, represent high statistical leverage of the $m^{th}$ negative bag's feature embeddings. Finally, we construct a key matrix $K$ of representative negative instances by concatenating $\tilde{A}_{b_m}$ for all $m = 1, 2, ..., P$,

$$K = \tilde{A}_{b_1} \oplus \tilde{A}_{b_2} ... \oplus \tilde{A}_{b_P} = [k_1, ..., k_\tau] \in \mathbb{R}^{D \times \tau} \quad (3)$$

where $\oplus$ denotes the concatenation operation and $\tau = \sum_{m=1}^{P} t_m$. This process is depicted in Algorithm 1.

---
**Algorithm 1.** Pseudo-code for negative representation learning (NRL)

**Input:** The set of embedded normal WSIs, $A = \{A_1, A_2, ..., A_P\}$.
**Step 1:** for $m = 1, 2, ..., P$ do

- **Step 1.1:** Compute the rank of $A_m$, and set $k = rank(A_m)$.
- **Step 1.2:** Construct matrix $V$ whose rows are the eigenvector of $A_m^T A_m$.
- **Step 1.3:** Compute the importance score of the $j^{th}$ column of $A_m$ by $s_j = \frac{1}{k}\sum_{h=1}^{k} V_{hj}^2$, where $V_{hj}$ is the element in the $h^{th}$ row and $j^{th}$ column of $V$, for $j = 1, 2, \ldots, n_m$.
- **Step 1.4:** Sort columns of $A_{b_m}$ based on the scores $s_j$'s.
- **Step 1.5:** Construct $\tilde{A}_{b_m} \in \mathbb{R}^{D \times t_i}$ whose columns are the first $t_m$ columns of sorted $A_{b_m}$ in Step 1.4.

**End (for).**

**Output:** Construct a key matrix $K$ by concatenating all $\tilde{A}_{b_m}$, for $m = 1, 2, \ldots, P$,
$$K = \tilde{A}_{b_1} \oplus \tilde{A}_{b_2} \ldots \oplus \tilde{A}_{b_P} = [k_1, \ldots, k_\tau] \in \mathbb{R}^{D \times \tau}$$

*2) Cross-attention-based salient instance inference MIL*

In this section, we introduce a new cross-attention-based salient instance inference MIL (CASiiMIL) model that can efficiently highlight salient instances of positive bags. Unlike existing attention-based MIL methods [2, 8, 11, 12], which learn attention weights solely from input instances, our cross-attention-based architecture can automatically correlate the input instances and negative keys, enabling the learning of high attention weights for instances that have low semantic relevance to normal tissues.

Suppose a fixed key matrix $K = [k_1, k_2, \ldots, k_\tau] \in \mathbb{R}^{D \times \tau}$ is constructed from all negative bags (see Section 3.3.1), and a random input bag (WSI) containing $n$ instances, $Q = [q_1, q_2, \ldots, q_n] \in \mathbb{R}^{D \times n}$, is given. The keys $k_1, k_2, \ldots, k_\tau$, and the queries $q_1, q_2, \ldots, q_n$ are first transformed into three latent feature spaces: key, query and value spaces, via three different fully connection layers:

$$\tilde{k}_j = FC^k(k_j) = \tanh(W_k^T k_j + b_k) \in \mathbb{R}^{D_h}, \quad (4)$$
$$\tilde{q}_i = FC^q(q_i) = \tanh(W_q^T q_i + b_q) \in \mathbb{R}^{D_h}, \quad (5)$$
$$\tilde{v}_i = FC^v(q_i) = \text{ReLU}(W_v^T q_i + b_v) \in \mathbb{R}^{D_h}, \quad (6)$$

where $W_k, W_q, W_v \in \mathbb{R}^{D \times D_h}$, and $b_k, b_q, b_v \in \mathbb{R}^{D_h}$ are learnable parameters. We then construct a cross-attention matrix $C$,

$$C = [c_{ij}] = \tilde{Q}^T \tilde{K} \in \mathbb{R}^{n \times \tau}, \quad (7)$$

where $\tilde{K} = \left[\frac{\tilde{k}_1}{||\tilde{k}_1||}, \frac{\tilde{k}_2}{||\tilde{k}_2||}, \ldots, \frac{\tilde{k}_\tau}{||\tilde{k}_\tau||}\right] \in \mathbb{R}^{D_h \times \tau}$ and $\tilde{Q} = \left[\frac{\tilde{q}_1}{||\tilde{q}_1||}, \frac{\tilde{q}_2}{||\tilde{q}_2||}, \ldots, \frac{\tilde{q}_n}{||\tilde{q}_n||}\right] \in \mathbb{R}^{D_h \times n}$. In matrix $C$, each row is a correlation vector between a query, $q_i$, to all the keys, $k_1, k_2, \ldots, k_\tau$.

Then, we construct a saliency layer, $FC^s$, whose inputs are rows of cross-attention matrix, $C_i \in \mathbb{R}^\tau$, and outputs are the saliency logits, $s_i \in \mathbb{R}$, for the queries:

$$s_i = FC^s(C_i) = W_s^T C_i + b_s \in \mathbb{R}. \quad (8)$$

Here, $W_s \in \mathbb{R}^{\tau \times 1}$ and $b_s \in \mathbb{R}$ are learnable parameters.

Finally, we compute a bag-level embedding for the input bag by aggregating the latent queries in value space as follows:

$$z = \sum_i^n a_i \tilde{v}_i, \quad (9)$$

where:
$$a_i = \frac{\exp(s_i)}{\sum_i^n \exp(s_i)}, \quad (10)$$

where $a_i$ are the saliency informed attention weights. Finally, we feed the bag-level embedding $z$ into a fully connected layer to classify the bag-level label (i.e., normal or tumor WSI) of the input bag in a supervised manner. The overall CASiiMIL architecture is depicted in Fig 1b.

### D. Model training

The proposed model is primarily trained with a binary cross-entropy loss function:
$$L_{CE} = -Y \log(P_{bag}) - (1-Y) \log(1 - P_{bag}), \quad (11)$$

where $Y \in \{0, 1\}$ is bag-level (slide-level) label of a WSI, and $P_{bag} \in [0, 1]$ is the bag-level probability of being positive predicted by CASiiMIL.

Moreover, we introduce two instance-level loss functions for the instances with bottom-r and top-r attention weights within each WSI. We propose these two instance-level loss functions to encourage the sparsity of the attention weights and guide CASiiMIL model to learn appropriate attention weights for each WSIs. The loss functions are as follows:

$$L_{bot} = -(1 - Y_0) \log(1 - P_s^{n-r,\ldots,n}), \quad (12)$$
$$L_{top} = -Y_1 \log(P_s^{1,\ldots,r}), \quad (13)$$

where:
$$P_s = \sigma(s), \quad (14)$$

where $Y_0 = 0$ and $Y_1 = 1$ are the pseudo-labels assigned to the bottom- and top-r instances, $s$ is the saliency logit (see Eq. 8) of one of the bottom or top-r instances, and $\sigma(\cdot)$ is the sigmoid activation function. In practice, $L_{bot}$ is applied for all training WSIs and $L_{top}$ is applied only for positive training WSIs. We train the model solely via $L_{CE}$ for 5 epochs to allow the model to warm-up. Each loss is averaged across the bottom- or top-r instances before optimization.

As a result, the proposed model is trained via:
$$L = L_{CE} + \lambda_1 L_{bot} + \lambda_2 L_{top}, \quad (15)$$

where $\lambda_1$ and $\lambda_2$ are constant coefficients for the instance-level losses.

### E. Histopathology specific feature extractor

In this study, we apply two different pretrained CNN feature extractors (i.e. $f(\cdot)$), namely ResNet50 [26] and CTransPath [27], for patch feature extraction. ResNet50 (truncated at the third residue block) is the most common feature extractor for MIL-based WSI analysis studies and pretrained on the ImageNet dataset [28]. The dataset contains more than one million natural images divided into 1000 categories, and this model has an output dimension of $D = 1024$. Despite its widespread use and successful applications [8, 12, 29, 30], the domain shift issue between natural images and histopathology images remains. Thus, we propose to use CTransPath, which is a transformer-based histopathology specific feature extraction model [27]. CTransPath is pretrained in a self-supervised learning manner using around 15 million histopathology image patches collected from the cancer genome atlas (TCGA) and pathology AI platform (PAIP) datasets. Its output dimension is $D = 768$.

## F. Implementation details

For preprocessing, all WSIs were cropped into patches in the size of 224×224 under 20× magnification (same as the settings of some recent studies [8, 11]). Patches from foreground tissue regions were extracted using color thresholding.

Our model was optimized by Adam optimizer [31] with 0.0002 learning rate and 0.00001 weight decay. The training was carried out with 10 epochs warm-up steps and halted if the validation AUC didn't improve for over 10 epochs. For the instance-level loss functions (Eq. 14 and 15), we empirically chose instances with bottom and top-5 attention weights since some WSIs have small tissue regions or micro-metastasis. Our code is available at: https://github.com/JoeSu666/CASiiMIL.

## G. Experimental design

To evaluate the performance of the proposed model, we run our model five times where we randomly split the training set of Camelyon16 into training and validation sets in a ratio of 9:1. Then, we selected the model with the best validation AUC from the five models and tested the model on the official testing set of Camelyon16.

To evaluate the cross-center generalizability of the proposed model, we also conducted five-fold cross-center cross-validation on the Camelyon17 dataset. Namely, in each fold, we employed WSIs from one center as the testing set, and combined the WSIs from the rest four centers and the Camelyon16 training WSIs as the training set. The training set was also randomly split into training and validation set in a ratio of 9:1. For comparison, we conducted the same experiments on the state-of-the-arts MIL methods[11, 32] [5, 12]. These two methods represent two different types of MIL frameworks: attention-based and self-attention-based methods. Both these methods have been reported to outperform other MIL frameworks. The experimental results are reported in Section 4.

## IV. RESULTS

### A. Slide-level classification

TABLE I

Slide-level classification results on Camelyon16 based on CTransPath feature extractor. 95% CI reported in [].

|  | AUC | F1 | PRECISION | RECALL |
|---|---|---|---|---|
| CLAM [8] | 0.9227 | 0.8913 | 0.9535 | 0.8367 |
|  | [0.9187, 0.9248] | [0.8838, 0.8967] | [0.9496, 0.9559] | [0.8296, 0.9400] |
| TransMIL [12] | 0.9313 | 0.8541 | 0.8723 | 0.8367 |
|  | [0.9286, 0.9331] | [0.8470, 0.8553] | [0.8667, 0.8745] | [0.8310, 0.8398] |
| DTFD-MIL [32] | 0.9408 | 0.8444 | 0.9268 | 0.7755 |
|  | [0.9382, 0.9422] | [0.8363, 0.8445] | [0.9238, 0.9313] | [0.7690, 0.7799] |
| CASiiMIL | **0.9679** | **0.9149** | **0.9556** | **0.8776** |
|  | **[0.9663, 0.9688]** | **[0.9101, 0.9151]** | **[0.9528, 0.9578]** | **[0.8722, 0.8799]** |

TABLE II

Slide-level classification results on Camelyon16 based on ResNet feature extractor. 95% CI reported in []

|  | AUC | F1 | PRECISION | RECALL |
|---|---|---|---|---|
| DSMIL [11] | 0.8265 | 0.6197 | **1.0000** | 0.4490 |
|  | [0.8219, 0.8281] | [0.6086, 0.6193] | [1.0000, 1.0000] | [0.4429, 0.4540] |
| CLAM [8] | 0.8319 | 0.7500 | 0.8462 | 0.6735 |
|  | [0.8298, 0.8367] | [0.7440, 0.7524] | [0.8380, 0.8475] | [0.6682, 0.6791] |
| TransMIL [12] | 0.8520 | 0.7907 | 0.9189 | 0.6939 |
|  | [0.8463, 0.8537] | [0.7867, 0.7946] | [0.9187, 0.9262] | [0.6915, 0.7022] |
| DTFD-MIL [32] | 0.8452 | 0.7579 | 0.7826 | 0.7334 |
|  | [0.8439, 0.8513] | [0.7508, 0.7598] | [0.7783, 0.7897] | [0.7311, 0.7425] |
| CASiiMIL | **0.8842** | **0.8222** | 0.9024 | **0.7551** |
|  | **[0.8804, 0.8860]** | **[0.8133, 0.8205]** | [0.8960, 0.9036] | **[0.7473, 0.7571]** |

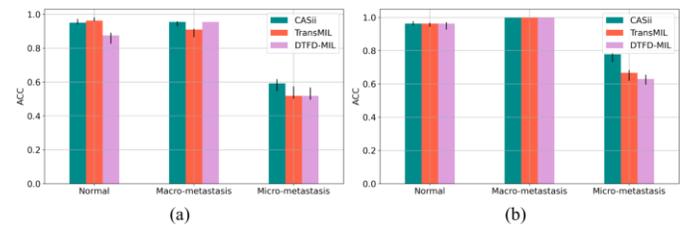

Fig 2. Slide-level classification accuracies with 95% CIs on Camelyon16 dataset that grouped in normal, macro-metastasis, and micro-metastasis. (a) Results based on CTransPath feature extractor. (b) Results based on ResNet50 feature extractor

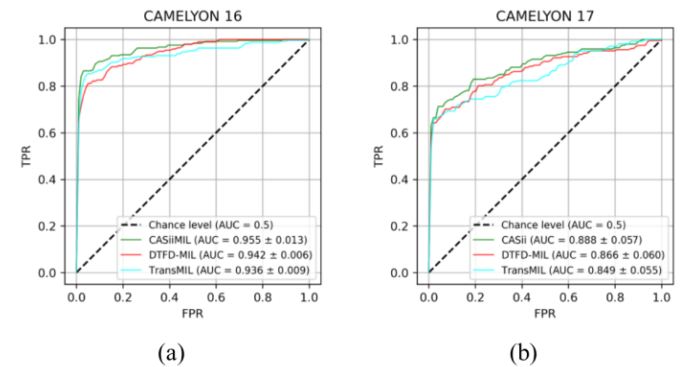

Fig 3. (a) Averaged ROC curve of the five runnings on Camelyon16 dataset. (b) Averaged ROC curve of five-fold cross-center cross-validation on Camelyon17 dataset. All results are based on CTransPath feature extractor.

Table I, Table II and Fig 2 report the slide-level classification results on the Camelyon16 dataset based on two different feature extractors. In comparison with state-of-the-art MIL models, CASiiMIL achieves the best overall performance. Unlike other methods, CASiiMIL maintains a good balance between precision and recall. Fig 2 reports the slide-level classification accuracies on the WSIs grouped in normal, macro-metastasis, and micro-metastasis. Moreover, Fig 3a reports the averaged ROC curve of the five runnings based on the Camelyon16 dataset.

TABLE III
Cross-center slide-level classification results on Camelyon17 dataset based on CTransPath feature extractor. 95% CI reported in [].

|  | Center0 | | Center1 | | Center2 | | Center3 | | Center4 | |
|---|---|---|---|---|---|---|---|---|---|---|
|  | AUC | F1 | AUC | F1 | AUC | F1 | AUC | F1 | AUC | F1 |
| TransMIL [12] | 0.8251 [0.8184, 0.8257] | 0.6415 [0.6308, 0.6434] | 0.7672 [0.7656, 0.7733] | 0.6250 [0.6179, 0.6278] | 0.8453 [0.8448, 0.8538] | 0.6552 [0.6484, 0.6598] | 0.9098 [0.9074, 0.9122] | 0.8108 [0.8021, 0.8103] | 0.9162 [0.9116, 0.9168] | 0.8696 [0.8679, 0.8746] |
| CLAM [8] | 0.8316 [0.8275, 0.8348] | 0.7018 [0.6928, 0.7018] | **0.8239** **[0.8221, 0.8279]** | 0.6290 [0.6237, 0.6301] | 0.9131 [0.9100, 0.9166] | 0.6667 [0.6537, 0.6683] | 0.9258 [0.9234, 0.9275] | 0.8421 [0.8357, 0.8428] | 0.9088 [0.9085, 0.9130] | 0.7792 [0.7742, 0.7824] |
| DTFD-MIL [32] | 0.8468 [0.8414, 0.8490] | **0.7500** **[0.7398, 0.7498]** | 0.7674 [0.7639, 0.7720] | 0.6667 [0.6572, 0.6667] | 0.8811 [0.8769, 0.8838] | 0.6897 [0.6778, 0.6892] | 0.9379 [0.9348, 0.9394] | **0.9041** **[0.8975, 0.9034]** | 0.9138 [0.9117, 0.9169] | 0.8732 [0.8667, 0.8733] |
| CASiiMIL | **0.8529** **[0.8498, 0.8567]** | 0.7368 [0.7281, 0.7370] | 0.7976 [0.7915, 0.7991] | **0.7000** **[0.6927, 0.7017]** | **0.9296** **[0.9248, 0.9302]** | **0.7600** **[0.7527, 0.7639]** | **0.9446** **[0.9422, 0.9457]** | 0.8493 [0.8439, 0.8509] | **0.9340** **[0.9317, 0.9365]** | **0.8889** **[0.8738, 0.8890]** |

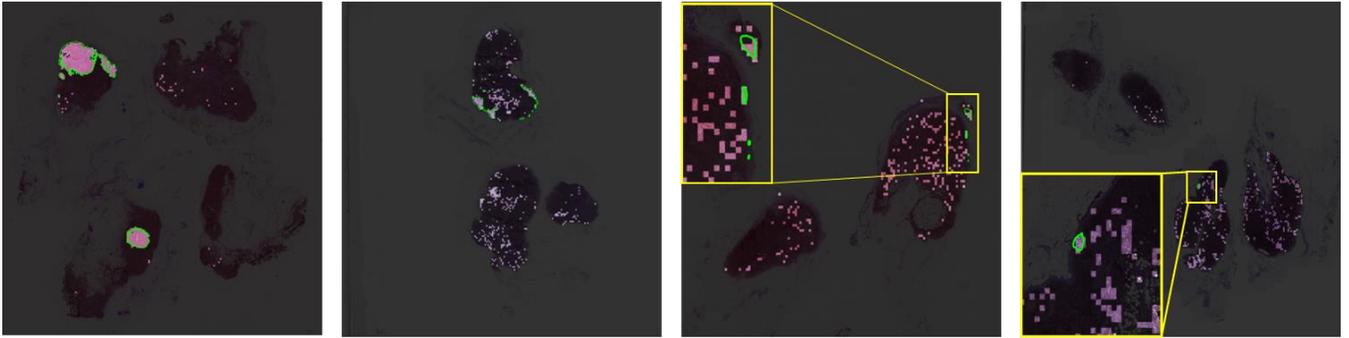

Fig 4. For attention visualization of example tumor WSIs, we highlight the patches that receive the largest 10% attention weights and dim (darken) the rest of the regions. Tumor regions are annotated in green color. From left to right, we visualize the example WSIs with different tumor sizes ranges from large to small. The results demonstrate the sensitivity of CASiiMIL on tumor lesions of different sizes.

Table III reports the cross-center slide-level classification results on the Camelyon17 dataset based on CTransPath feature extractor. In comparison with state-of-the-art MIL models, CASiiMIL exhibits better cross-center generalizability.

### B. Ablation studies

To demonstrate the effect of different loss functions, we conduct an ablation study on different combination of our loss function terms. The results are summarized in Table 4.

TABLE IV
Ablation study on loss functions. Results are averaged testing results on Camelyon16 dataset

|  | AUC | F1 |
|---|---|---|
| $L_{CE}$ | 0.9456 | 0.8736 |
| $L_{CE} + L_{bot}$ | 0.9514 | 0.8657 |
| $L_{CE} + L_{top}$ | 0.9527 | 0.8855 |
| $L_{CE} + L_{bot} + L_{top}$ | **0.9545** | **0.8919** |

### C. Interpretability

To demonstrate the interpretability of the proposed model, we visualize the attention outputs (see Eq. 10) of CASiiMIL overlaid on the original WSIs in Fig 4. Specifically, we highlight the patches that receive the largest 10% attention weights and dim the remaining regions. The visualization results demonstrate CASiiMIL's sensitivity on tumor lesions in different sizes. Additionally, we invited a pathologist to interpret the false positive patches that received the largest 10% attention weights but were not from the annotated tumor regions, according to Camelyon16's official annotation. We find that most false positive patches are normal cells such as histiocytes (tissue macrophages) and high endothelia cells that shared similar morphology with tumor cells, such as large cell size, large but less blue nucleus, profound nucleolus, etc. Some example patches are shown in Fig 5.

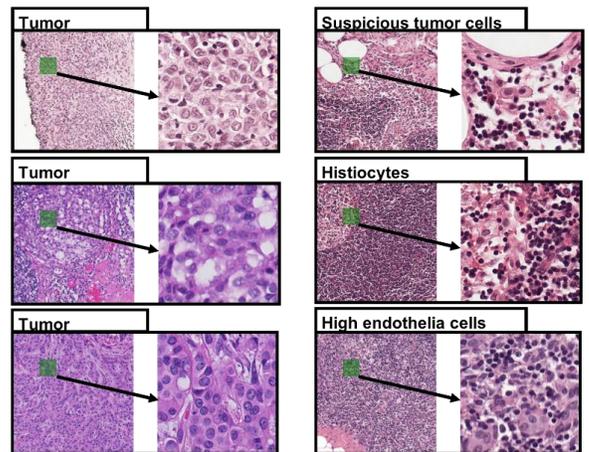

Fig 5. Interpretation of patches with high attention weights. Here, we show example patches that receive the largest 10% attentions weights within their corresponding WSIs. (a) True positive patches that are from annotated tumor regions. (b) False positive patches that are not from annotated tumor regions.

## V. Discussions and conclusions

This manuscript makes a significant contribution to the field of pathology image analysis by proposing a novel method for automatically computing a saliency score for each patch in an image, which can be used to arrive at a slide-level decision. This method has several advantages over existing approaches, including the ability to produce more reliable and interpretable attention maps, as well as the ability to correctly identify cases with extremely small lesions that may be missed by other methods. By computing a saliency score for each patch, our method is able to identify regions of interest within an image that are most likely to contain pathological features. This allows us to focus our analysis on these regions and improve the accuracy of our predictions. Additionally, by combining these scores across all patches in an image, we are able to arrive at a slide-level decision that takes into account all available information. One of the key advantages of our method is its ability to identify cases with extremely small lesions accurately. This is particularly important in pathology image analysis, where small lesions can be easily missed or overlooked by human observers. By automatically computing saliency scores for each patch, our method is able to detect even very small lesions and incorporate this information into the slide-level decision. Overall, our method represents a significant advance in the field of pathology image analysis and has the potential to improve the accuracy and reliability of diagnostic and prognostic predictions.

We innovatively developed a cross-attention architecture to integrate a salient instance inference module into the gradient-flow of MIL classification. There are two main advantages associated with our model. First, it transforms the key and queries to a latent space to correlate query instances with the negative keys automatically. Second, it enables the learning of saliency informed attention weights to highlight the possible positive instances in the bags. In section 4.A, we show that the proposed CASiiMIL achieves outstanding slide-level classification performance on Camelyon16 dataset with both natural image pretrained and histopathology-specific feature extractors (see Table I and II). In addition, CASiiMIL exhibits the best recall without sacrificing precision, which is especially important for a diagnostic tool in clinical practice.

In Fig 2, we compare different MIL models' slide-level classification accuracies on the WSIs grouped in normal, macro-metastasis, and micro-metastasis. The proposed CASiiMIL achieves best accuracies on the micro-metastasis WSIs compared with all other MIL models. This result demonstrates that our cross-attention architecture and saliency informed attention weights are helpful on identify tumor WSI, even when the tumor lesions are small. In Table III, we report the cross-center slide-level classification results of CASiiMIL compared with other MIL methods. The results reveal that the proposed model can greatly generalize to WSIs from unseen hospitals during training.

In the visualizations of attention heatmaps (see Fig 4), we further show that our model is able to identify tumor lesions and predict high attention weights on the tumor instances. This outcome demonstrates the outstanding sensitivity and interpretability of the proposed saliency informed attention layer regardless of the size of tumor lesions. Besides tumor lesions, this model is also identifying several groups of benign cells (except for lymphocytes) that are normal components of lymph nodes. These cell groups are mostly high endothelial venules and histiocytes (tissue macrophages). They may share some similar morphological features with tumor cells in comparison to background lymphocytes, such as larger cell size, larger nucleus with profound nucleoli, and abundant cytoplasm. Notably, the model also identified a few isolated cells that were highly suspicious for tumor cells, based on the morphology comparing to the confirmed main tumor lesion in the same lymph node (see Fig 5).

A limitation of the proposed method is that CASiiMIL is computationally expensive due to feeding the entire key set into the model during the training. Approximately, our model has 893K learnable parameters and 6M non-learnable parameters (i.e., the entire negative key set). This manner requires key sets of limited size given certain computational memories. To this end, a more robust negative representation learning method is needed to extract a more representative and less redundant negative key set.

In summary, we propose a novel MIL model called CASiiMIL, which achieved excellent accuracy on the tumor WSI classification task. We innovatively developed cross-attention-based architecture that enables the learning of saliency informed attention weights for MIL aggregation. The proposed CASiiMIL is of great sensitivity and interpretability in classifying tumor WSIs regardless the how small the tumor regions are, which makes it a reliable automatic diagnostic tool.